\documentclass[runningheads]{llncs}
\usepackage[T1]{fontenc}
\usepackage{graphicx}
\usepackage{booktabs}
\usepackage[misc]{ifsym}
\newcommand{\corr}{(\Letter)}
\usepackage{amsmath}
\usepackage{multirow}
\usepackage{subcaption}
\usepackage{hyperref}

\begin{document}

\title{Debate-on-Graph: Reliable and Adaptive Reasoning of Large Language Model on Uncertain Knowledge Graph}

\titlerunning{Debate-on-Graph}

\author{Peiji Yu \inst{1} \and
Xin Chen \inst{1} \and
Tianxing Wu \inst{1,2}\footnote[1]{Corresponding author.} \corr}

\authorrunning{P. Yu et al.}

\institute{School of Computer Science and Engineering, Southeast University, Nanjing, China \email{\{213232231,213231269,tianxingwu\}@seu.edu.cn}
\and
Key Laboratory of New Generation Artificial Intelligence Technology and Its Interdisciplinary Applications (Southeast University), Ministry of Education, China}

\tocauthor{Peiji Yu, Xin Chen, Tianxing Wu}
\toctitle{Debate-on-Graph: Reliable and Adaptive Reasoning of Large Language Model on Uncertain Knowledge Graph}

\maketitle              

\begin{abstract}
Large language models (LLMs) have demonstrated remarkable capabilities in natural language processing. However, LLMs often suffer from hallucinations and lack of relevant knowledge when dealing with question answering (QA) tasks. To mitigate these issues, knowledge graphs (KGs) have been utilized to enhance LLM reasoning. Nevertheless, KGs often contain noise and errors, while existing KG-enhanced LLM approaches are generally unable to identify and filter such noisy and erroneous content, which can instead amplify hallucinations and pose challenges for reliable reasoning. Uncertain knowledge graphs (UKGs), which associate each triple with a confidence score to quantify uncertainty, offer a promising direction to address this challenge. Compared with prior work, we investigate how to leverage UKGs to support LLMs for QA. We propose \textbf{D}ebate-\textbf{o}n-\textbf{G}raph (\textbf{DoG}), a new framework that enables LLMs and UKGs to collaborate adaptively for reliable reasoning. Specifically, we first design a heuristic search algorithm tailored for UKGs to extract reliable and question-relevant subgraphs, thereby reducing noise and errors in retrieved knowledge. We then introduce a Multi-Agent Debate mechanism, which yields reliable answers through adaptive adversarial debates, aiming to fully exploit the knowledge in UKGs while preserving the reliability of retrieved evidence. Extensive experiments on four benchmark QA datasets show that DoG achieves state-of-the-art performance over existing LLM reasoning methods and KG-based baselines, while enabling reliable and adaptive reasoning. Our code is available at \href{https://github.com/seucoin/Debate-on-Graph}{https://github.com/seucoin/Debate-on-Graph}.

\end{abstract}

\section{Introduction}
Large language models (LLMs) have demonstrated remarkable capabilities in natural language processing \cite{brown2020language,hurst2024gpt,team2023gemini,yang2025qwen3}. However, when dealing with question answering (QA) tasks, LLMs often suffer from hallucinations \cite{ji2023survey} and lack of relevant knowledge, which can undermine the reliability of their responses. To mitigate these issues, knowledge graphs (KGs) \cite{hogan2021knowledge}, which describe real-world facts in the form of triples denoted as $(head, relation, tail)$, or simply $(h, r, \tau)$, have been frequently utilized to enhance LLM reasoning \cite{sunthink,luo2024reasoning}. As a structured and non-parametric external knowledge source, KGs offer a complementary strategy to alleviate the inherent limitations of LLMs \cite{ma2025large,pan2024unifying}.

Prior studies on KG-enhanced LLM reasoning can be broadly categorized into two paradigms: \textit{retrieval-based} and \textit{agent-based} approaches \cite{luo2024graph}. Retrieval-based approaches \cite{luo2024reasoning,wen2024mindmap,mavromatis2024gnn} employ an external retriever to fetch relevant facts from KGs and incorporate them into the LLM input for reasoning, while agent-based approaches \cite{sunthink,tan2025paths,fang2025karpa} treat LLMs as agents that directly interact with KGs, starting from an initial entity and iteratively exploiting reasoning paths until the LLM decides the augmented knowledge is sufficient.

However, since the automatic process has been widely applied in the construction of KGs, they may often contain noise and factual errors \cite{zhang2021gaussian}. Despite notable progress, both categories of KG-enhanced approaches remain sensitive to such imperfections, posing challenges for reliable reasoning. Specifically, in retrieval-based approaches, retrievers may retrieve irrelevant or incorrect facts, which can mislead generation; in this case, the injected evidence not only fails to reduce hallucinations but may even amplify them \cite{mavromatis2024gnn}. For agent-based approaches, noisy or erroneous triples can mislead the LLM at early interaction steps, making the entire multi-step process unreliable \cite{sunthink}.

Uncertain knowledge graphs (UKGs), which have attracted increasing attention recently, are ideally suited for capturing the uncertainties inherent in real-world scenarios. A UKG associates each triple with a confidence score to quantify uncertainty, where the score represents the probability of that triple being true \cite{carlson2010toward,speer2017conceptnet}. Thus, each fact is expanded from a triple $(h, r, \tau)$ to a quadruple as $(h, r, \tau, c)$. For example, \textit{(Twitter, competeswith, Facebook, 0.85)} represents the probability of the fact “Twitter competes with Facebook” is 0.85. This design captures the inherent uncertainty of real-world knowledge, which paves the way for identifying and filtering noisy or erroneous knowledge, opening up new opportunities for reliable reasoning.

Compared with prior work, we investigate how to leverage UKGs to assist LLMs for QA (i.e., the UKGQA task) for reliable reasoning. Here, a key new challenge is \textbf{how to fully exploit the knowledge in UKGs while reserving the reliability of retrieved evidence}, as there is always a trade-off between knowledge utility and reliability. We propose that an ideal approach should adapt to each question, striking a balance between knowledge utility and reliability, and using this balance to guide retrieval so that LLMs can reduce hallucinations and perform reliable and effective reasoning.

To address this challenge, we propose \textbf{D}ebate-\textbf{o}n-\textbf{G}raph (DoG), a new framework that enables LLMs and UKGs to collaborate adaptively for
reliable reasoning. Specifically, we first design a heuristic search algorithm for UKGs to extract reliable and question-relevant subgraphs, thereby reducing noise and errors in retrieved knowledge. The algorithm also supports tunable hyperparameter to dynamically expand the subgraph if needed. We then introduce a \textbf{Multi-Agent Debate (MAD)} mechanism \cite{du2023improving,hu2025removal} with four roles: a Proponent Agent, a Challenger Agent, a Decision-Making Agent, and a Judge Agent. The Proponent and Challenger engage in adversarial debates under an information-asymmetric setting, which reduces the LLM’s over-reliance on retrieved content and improves robustness to noisy or erroneous facts. The Decision-Making Agent subsequently determines whether expand the subgraph or terminate the debate, aiming to fully exploit the UKG while keeping retrieval reliable. Finally, the Judge Agent synthesizes the debate outcomes and determines the final answer.

Overall, the contributions of our work are three-fold:
\begin{itemize}
    \item[$-$] We present Debate-on-Graph, a new framework that facilitates adaptive collaboration between LLMs and UKGs for reliable reasoning.
    \item[$-$] We propose a heuristic search algorithm tailored for UKGs that can dynamically extract reliable and question-relevant subgraphs, thereby reducing noise and errors in the retrieved knowledge.
    \item[$-$] We design a MAD-based answer generation strategy that guides LLMs to yield reliable answers through adaptive interaction with UKG subgraphs, aiming to fully exploit the UKG while keeping retrieval reliable.
\end{itemize}

\section{Related Work}

\subsection{Uncertain Knowledge Graph}
Deterministic Knowledge Graphs (DKGs), such as Freebase \cite{bollacker2008freebase} and Wikidata \cite{vrandevcic2014wikidata}, consist of deterministic facts, have been utilized for different applications \cite{guo2020survey,chen2022outlining,stathopoulos2024applied}. In contrast, UKGs associate each triple with a confidence score, thereby capturing the universal uncertainty and fuzziness of real-world knowledge. Representative examples of UKGs include NELL \cite{carlson2010toward} and ConceptNet \cite{speer2017conceptnet}.

Recently, the task of UKG completion, which encompasses confidence prediction and link prediction, has attracted increasing attention \cite{wang2024unkr}. Current research \cite{chen2019embedding,chen2021passleaf,yang2022approximate,chen2024uncertain,wu2026uncertain} in this domain primarily focuses on normal relational learning, which embeds entities and relations while preserving both graph structures and confidence information. Moreover, recent studies have extended this focus to few-shot relational learning to address the long-tail distribution of relations commonly observed in real-world UKGs \cite{zhang2021gaussian,wang2022incorporating}. However, the task of leveraging UKG to support LLMs for QA remains unexplored. UKGs are essential for many downstream applications, such as dietary recommendation \cite{sha2025leveraging} and upper limb motor recovery \cite{wu2025supervised}.

\subsection{KG-enhanced LLM Reasoning}
To mitigate the hallucinations and knowledge gaps inherent in LLMs, integrating KGs with LLMs has established itself as a critical research direction \cite{pan2024unifying,WOS:001356154700001}. Existing methodologies can be primarily categorized into \textit{retrieval-based} and \textit{agent-based} paradigms.

Retrieval-based approaches focus on identifying and extracting relevant knowledge from KGs to augment the input context of the LLM. Early works in this domain often linearize knowledge triples directly, whereas recent advancements exploit the structural information of the graph. For instance, MindMap \cite{wen2024mindmap} utilizes KGs to elicit a cognitive map for the LLM, thereby revealing reasoning pathways. RoG \cite{luo2024reasoning} generates relation paths as plans to retrieve valid reasoning subgraphs. Similarly, GNN-RAG \cite{mavromatis2024gnn} employs Graph Neural Networks to retrieve answer candidates and extract the shortest paths connecting question entities. While effective for direct fact retrieval, these approaches often overlook the rich structural dependencies within KGs and struggle to handle the noise introduced by irrelevant retrieved information.

In contrast, agent-based approaches treat the LLM as an autonomous agent that interactively explores the KG to perform complex reasoning. Instead of a static retrieval step, these approaches navigate the graph structure iteratively to locate answers. For instance, ToG \cite{sunthink} implements a beam search strategy, enabling the LLM to discover reasoning paths step-by-step. PoG \cite{tan2025paths} focuses on pruning irrelevant information to enhance the faithfulness and interpretability of the reasoning process. KARPA \cite{fang2025karpa} leverages the global planning capabilities of LLMs to pre-plan relation paths, matches them via an embedding model, and aggregates them to avoid stepwise local optima. However, agent-based paradigms are prune to errors, as noisy or erroneous triples encountered during interaction can significantly mislead the entire trajectory of the LLM on KGs.

\section{Preliminary}

In this section, we introduce the basic concepts and notation used throughout the paper.

\begin{definition}[\textbf{Uncertain Knowledge Graph}]
An Uncertain Knowledge Graph is defined as $\mathcal{G} (\mathcal{E}, \mathcal{R}, \mathcal{T})$, where $\mathcal{E}$, $\mathcal{R}$ and $\mathcal{T}$ denote the set of entities, relations and uncertain knowledge triples, respectively.
\end{definition}

\noindent Each triple $t \in \mathcal{T}$ is a quadruple:
\begin{equation}
t = (h, r, \tau, c), \quad h,\tau \in \mathcal{E},\ r \in \mathcal{R},\ c \in (0,1],
\end{equation}
where $c$ is the confidence score indicating the probability that the triple is true.

\begin{definition}[\textbf{Uncertain Knowledge Graph Question Answering}]
Given a natural language question $Q$ and a UKG $\mathcal{G}$, Uncertain Knowledge Graph Question Answering is a task aiming to design a function $f$ that predicts an answer $a$ using the knowledge from $\mathcal{G}$, i.e., $a = f(Q, \mathcal{G})$.
\end{definition}

\begin{definition}[\textbf{Topic Entities}]
Topic Entities are a set of entities: $\mathcal{E}_{\text{topic}}=\{e_1,e_2,\ldots,e_J\}$ that are related to the UKG-based question, where $e_j \in \mathcal{E}$ denotes the $j$-th entity in the question $Q$. 
\end{definition}

\noindent The set of topic entities can be obtained via named entity recognition (NER) and entity linking techniques.

\begin{definition}[\textbf{Reasoning Paths}]
A Reasoning Path from entity $u$ to entity $v$ is an ordered sequence of triples:
\begin{equation}
p = \langle t_1, t_2, \ldots, t_k \rangle,
\end{equation}
where each triple has the form $t_i=(h_i,r_i,\tau_i,c_i)$ and $h_1=u,\quad \tau_k=v,\quad \tau_i = h_{i+1}\ \text{for}\ i=1,\ldots,k-1.$
\end{definition}



\begin{definition}[\textbf{Path Confidence}]
Given a reasoning path $p=\langle t_1,\ldots,t_k\rangle$ with triple confidences $\{c_i\}_{i=1}^k$, the confidence of the whole path is defined as the product of its triple confidences:
\begin{equation}
Conf(p)=\prod_{i=1}^{k} c_i.
\end{equation}
\end{definition}

\begin{definition}[\textbf{Path–Question Semantic Relevance}]
Let $f(\cdot,\cdot)$ denote a similarity function between the path $p$ and the question $Q$:
\begin{equation}
SR(p,Q)=f(p,Q).
\end{equation}
\end{definition}

\noindent We calculate $f(\cdot,\cdot)$ using a pre-trained language model (PLM), e.g. SentenceBERT \cite{reimers2019sentence}.

\begin{definition}[\textbf{Reasoning Path Score}]
Given a question $Q$, the Reasoning Path Score of path $p$ is determined by their semantic relevance and Path Confidence:
\begin{equation}
Score(p,Q)=SR(p,Q)\cdot (Conf(p))^{\alpha}
\label{score}
\end{equation}
where $\alpha$ is a hyperparameter controlling the trade-off between question relevance and Path Confidence.
\end{definition}

\noindent This matches the intuition that a path should be preferred when it is both semantically aligned with the question and supported by high-confidence triples.

\begin{definition}[\textbf{Entity Distance}]
The entity distance between entities $u$ and $v$ is defined as:
\begin{equation}
Distance(u,v)=\min_{p\in \mathcal{P}(u,v)} \frac{Hop(p)}{Conf(p)}.
\end{equation}
where $\mathcal{P}(u,v)$ denotes the set of reasoning paths from $u$ to $v$, and $Hop(p)$ represents the number of hops in $p$.
\end{definition}

\noindent Entity Distance favors shortness (as smaller $Hop(p)$ reduces the distance) and reliability (as larger $Conf(p)$ reduces the distance), thereby prioritizing paths that are both concise and trustworthy.

\section{The Approach: Debate-on-Graph}
\begin{figure}[h]
    \centering
    \includegraphics[width=\linewidth]{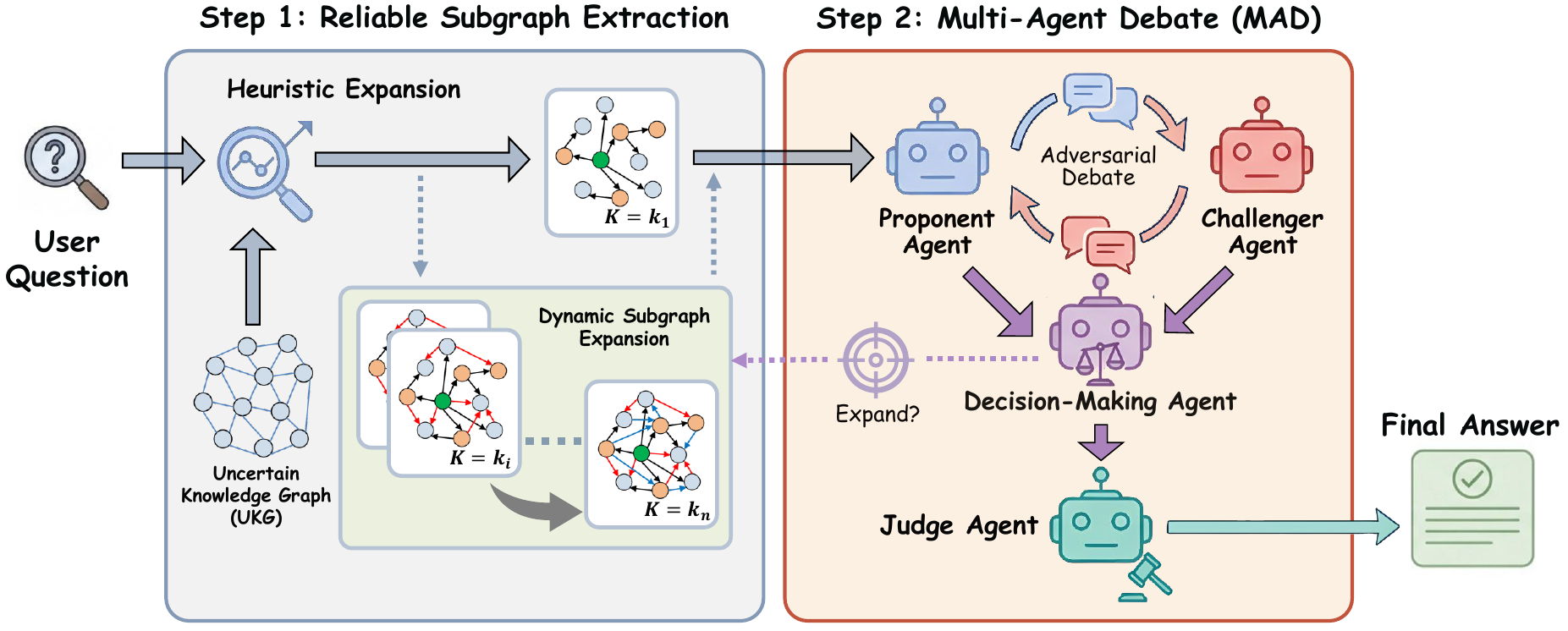}
    \caption{The overview of Debate-on-Graph framework.}
    \label{fig:overview}
\end{figure}

The overall framework of DoG, as illustrated in Fig. \ref{fig:overview}, consists of two steps: Step 1 for reliable subgraph extraction and Step 2 for multi-agent based answer generation, which are detailed in Sections \ref{subgraph} and \ref{MAD}, respectively.

For UKGQA tasks, given a question $Q$, we first obtain the set of topic entities via NER and entity linking techniques. Subsequently, for each topic entity, we employ the subgraph extraction algorithm to extract a question-relevant subgraph\footnote[1]{For question with multiple topic entities, we simply perform subgraph extraction for each one of them, retrieving a corresponding set of subgraphs to answer the question.}. Finally, we convert the subgraph into textual descriptions, and apply the multi-agent based framework to obtain the final answer.

\subsection{Reliable Subgraph Extraction}\label{subgraph}

\begin{figure}[h]
    \centering
    \includegraphics[width=\linewidth]{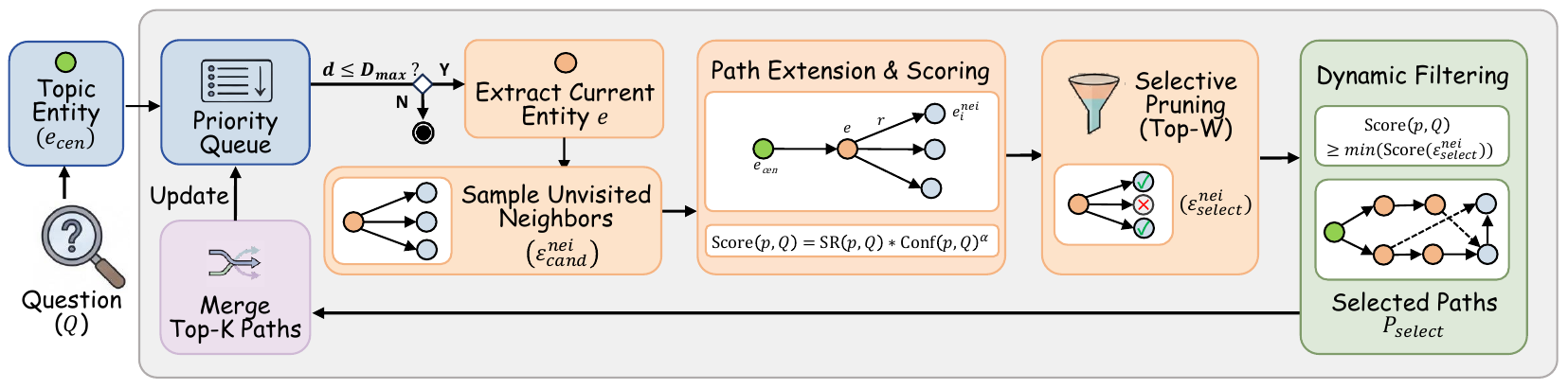}
    \caption{The overview of the subgraph extraction algorithm.}
    \label{fig:subgraph_extraction}
\end{figure}

To effectively reduce noise and errors in UKGs, we propose a heuristic subgraph extraction algorithm, as illustrated in Fig. \ref{fig:subgraph_extraction}. This algorithm operates on a best-first search paradigm, designed to dynamically extract question-relevant subgraphs that are both semantically aligned with the query and factually reliable. The algorithm initiates with a topic entity $e_{cen} \in \mathcal{E}_{topic}$, which serves as the central entity and a potential starting anchor for the search process. To manage the search space efficiently, we maintain a priority queue that orders candidate entities based on their distance to the center entity.

For a specific entity $e$ in UKG $\mathcal{G}$, the algorithm iteratively explores its neighborhood to identify the most promising expansion paths. Specifically, we first sample all unvisited direct neighbors $\mathcal{E}^{nei}_{cand} = \{e^{nei}_1, e^{nei}_2, \ldots, e^{nei}_n\}$ of $e$. The core challenge lies in evaluating the utility of these neighbors given the uncertainty inherent in the UKG. To address this, we retrieve the set of existing reasoning paths $\mathcal{P}_{cand}(e^{nei}_i)$ from $e_{cen}$ to $e$ and extend them by appending the connecting triple $(e, r, e_i^{nei})$ for each neighbor $e^{nei}_i$, generating a candidate set of extended paths defined as:
\begin{equation}
\mathcal{P}_{cand}(e^{nei}_i) = \{ p \oplus (e, r, e^{nei}_i) \mid p \in \mathcal{P}_{select}(e) \},
\end{equation}
where $\oplus$ denotes the path concatenation operation. We then evaluate each of these extended paths individually using the path scoring function $Score(p, Q)$ defined in Eq. (\ref{score}), which integrates both the semantic relevance $SR(p, Q)$ and the path confidence $Conf(p)$.


To prevent the retrieval of irrelevant noise and to control the scale of the subgraph, we employ a selective pruning strategy. Instead of expanding all neighbors, we select only the top-$W$ entities with the highest maximum path scores. Formally, we define the score of a neighbor $e_{nei}$ as the maximum score among all paths reaching it, and retrieve the set of selected entities $\mathcal{E}_{select}^{nei}$ as:
\begin{equation}
\mathcal{E}_{select}^{nei} = \mathop{\mathrm{Top\text{-}}W}\limits_{e_{nei} \in \mathcal{E}^{nei}_{cand}} \left( \max_{p \in \mathcal{P}_{cand}(e_{nei})} \mathrm{Score}(p, Q) \right)
\end{equation}
where $\mathcal{P}_{cand}(e_{nei})$ denotes the set of extended paths from the center entity $e_{cen}$ to neighbor $e_{nei}$. This hyperparameter $W$ acts as an adaptive expansion width, allowing the algorithm to focus computational resources on the most plausible branches of knowledge.


Furthermore, to ensure the quality of the retrieved evidence, we calculate a dynamic scoring threshold based on these selected entities. Only reasoning paths that exceed this threshold are retained in the path set $\mathcal{P}_{select}$ for each selected neighbor:
\begin{equation}
\mathcal{P}_{select} = \left\{ p \in \mathcal{P}_{cand} \mid Score(p, Q) \ge \min_{e \in \mathcal{E}_{select}^{nei}} Score(e_{nei}) \right\},
\end{equation}
where $\mathcal{P}_{cand}$ denotes the set of candidate paths extending to the selected entity and $Score(e_{nei})$ represents the score of entity $e_{nei}$.


Subsequently, the algorithm updates the priority queue with these selected neighbors. If a neighbor $e_{nei}$ is encountered for the first time, it is inserted into the queue with a priority determined by its Entity Distance to the center entity. Conversely, if $e_{nei}$ is already present in the queue, we merge the newly discovered paths with the existing ones, keeping only the top-$K$ paths with the highest scores, and update its priority if a shorter or more confident path is found. The expansion process repeats by extracting the entity with the minimum Entity Distance from the priority queue, until the distance of the closest candidate exceeds a pre-defined maximum threshold $D_{max}$. Finally, the extracted subgraph is constructed from the union of all high-quality triples contained within the reasoning path sets of the visited entities, providing a robust and concise knowledge basis for the subsequent reasoning phase.

\subsubsection{The critical role of $K$ in the algorithm}\label{K}
As a critical hyperparameter in the subgraph extraction algorithm, $K$ dictates the maximum number of reasoning paths retained for each entity. It effectively modulates the trade-off between knowledge utility and reliability. When $K$ is set to a conservative value, the constructed subgraph contains only the most reliable and semantically relevant paths; however, this strict filtering may result in an information deficit, preventing the model from deriving a determinate answer. Conversely, an excessively large $K$ incorporates a wealth of information but inevitably introduces lower-confidence paths and noise, thereby amplifying the risk of hallucinations and diminishing the reliability of the reasoning process. Consequently, finding an optimal $K$ is essential to balance knowledge utility against evidence reliability, which will be reflected in the next section.

Due to the space limitation, the pseudo-code of subgraph extraction is shown in the supplementary material.

\subsection{Multi-Agent Debate for Answer Generation}\label{MAD}

Despite the efficacy of the heuristic search algorithm in extracting relevant subgraphs, there may be still some residual noise or factual inaccuracies in the extracted context, which can inadvertently mislead the LLM \cite{hu2025removal}. Meanwhile, we aim to fully exploit the knowledge in UKGs while reserving the reliability of retrieved evidence. To tackle these challenges, we propose a \textbf{Multi-Agent Debate (MAD)} mechanism \cite{du2023improving,hu2025removal} that dynamically navigates the trade-off between knowledge utility and evidence reliability. This mechanism operates by adaptively adjusting the subgraph expansion parameter $K$ to guide the LLM toward a reliable answer.

The MAD-based framework consists of four distinct roles: a \textbf{Proponent Agent} $R_p$, a \textbf{Challenger Agent} $R_c$, a \textbf{Decision-Making Agent} $R_d$, and a \textbf{Judge Agent} $R_j$. A critical feature of this design is an information-asymmetric setting between the debating roles. The Proponent Agent is granted access to the retrieved UKG subgraph description corresponding to the current expansion level $K$, allowing it to ground its arguments in external evidence. In contrast, the Challenger Agent is restricted from accessing the subgraph, forcing it to rely exclusively on its internal knowledge. This design is critical to mitigating the LLM's tendency to over-rely on retrieved context; if both agents possessed identical access to the subgraph, they might unthinkingly converge on erroneous retrieved facts rather than critically evaluating their validity \cite{hu2025removal}.

The debate unfolds as an iterative optimization process over a pre-defined candidate set of expansion thresholds $\Theta = \{k_1, k_2, \dots, k_n\}$, $k_j > k_i$ if $j > i$. In the initial round, the Proponent generates a preliminary response based on the most reliable subgraph $\mathcal{G}_{Q}^{k_1}$ constructed with $K=k_1$, while the Challenger initiates the response based on its internal knowledge:
\begin{equation}
    y^1_{p} = \mathcal{R}_{p}(Q, \mathcal{G}_{Q}^{k_1})
\end{equation}
\begin{equation}
    y^1_{c} = \mathcal{R}_{c}(Q)
\end{equation}
where $y^{1}_{p}$  and $y^{1}_{c}$ represent the response from the Proponent Agent and Challenger Agent in the first round, respectively.

In subsequent rounds, both agents refine their responses by critiquing the opponent's previous response, thereby exposing inconsistencies or logical gaps:
\begin{equation}
    y^i_{p} = \mathcal{R}_{p}(Q, \mathcal{G}_{Q}^{k_j}, \{y^{<i}_{p}\}, y^{i-1}_{c}), \quad \text{for } i \geq 2
\end{equation}
\begin{equation}
    y^i_{c} = \mathcal{R}_{c}(Q, \{y^{<i}_{c}\}, y^{i-1}_{p}), \quad \text{for } i \geq 2
\end{equation}

Throughout this process, the Decision-Making Agent acts as a meta-controller, evaluating the state of the debate after every round to select one of three actions: First, if the debate is thorough and the debaters reach a consensus, the debate is terminated. Second, if the agents remain in disagreement but the evidence is sufficient, the debate proceeds to the next round to further clarify the arguments. Third, and most notably, if the Decision-Making Agent identifies that the Proponent lacks sufficient evidence to refute the Challenger or support its claim, it triggers an expansion of the subgraph from current level $k_j$ to the next level $k_{j+1}$ and continues the debate. This adaptive expansion, formalized as Eq. (\ref{action}), injects richer knowledge into the Proponent's context when necessary, allowing for more comprehensive reasoning in the subsequent round. In this way, DoG realizes adaptive reasoning by dynamically selecting both the evidence scope and the number of debate rounds for each question.
\begin{equation}
\label{action}
    \text{ACTION}_i = \mathcal{R}_d(Q, \mathcal{G}_{Q}^{k_j}, y^{i}_p, y^{i}_c)
\end{equation}
where $\text{ACTION}_i$ denotes one of three aforementioned actions.

Finally, once the debate concludes or reaches the maximum debate round $r_{max}$, the Judge Agent synthesizes the debate outcomes to derive a final, verified answer $a$.
\begin{equation}
    a = \mathcal{R}_j(Q, y^{r}_p, y^{r}_c)
\end{equation}
where $r \leq r_{max}$ denotes the final round of debate.

The overall Multi-Agent Debate procedure and the prompt templates for the debate agents can be found in the supplementary material.

\section{Experiments}
In the experiments, we aim to answer the following research
questions: \textbf{RQ1:} How does DoG perform compared with state-of-the-art LLM reasoning methods and KG-based baselines across different QA benchmarks? \textbf{RQ2:} What is the contribution of each component in DoG? \textbf{RQ3:} Can DoG perform adaptive reasoning? \textbf{RQ4:} How do hyperparameters affect the performance of DoG? \textbf{RQ5:} How efficient is DoG in terms of runtime, LLM calls, and token usage compared to other baselines?

\subsection{Experimental Setup}

\subsubsection{Datasets}
To comprehensively evaluate the effectiveness of our proposed framework, we conduct experiments on four benchmark datasets across two distinct domains: Commonsense QA and Medical QA. For the domain of commonsense QA, we select CommonsenseQA (CSQA) \cite{talmor2019commonsenseqa}, CommonsenseQA 2.0 (CSQA 2.0) \cite{talmor2022commonsenseqa}, and OpenBookQA (OBQA) \cite{OpenBookQA2018}, utilizing ConceptNet \cite{speer2017conceptnet} as the background UKG. For Medical QA, we employ MedQA-US, with MedKGent \cite{zhang2025medkgent} as the underlying UKG. The statistics of the datasets are summarized in Table \ref{tab:datasets}, while details are provided in the supplementary material.

\begin{table}[htbp]
\centering
\setlength{\tabcolsep}{5pt} 
\caption{The statistics of datasets and UKGs used in experiments.}
\label{tab:datasets}
\begin{tabular}{lcccc}
\toprule
\textbf{Dataset} & \textbf{Domain} & \textbf{Type} & \textbf{Size} & \textbf{UKG} \\
\midrule
CSQA & Commonsense & Multiple choice (5 options) & 500 & ConceptNet \\
CSQA 2.0 & Commonsense & True / False & 500 & ConceptNet \\
OBQA & Commonsense & Multiple choice (4 options) & 500 & ConceptNet \\
MedQA-US & Medical & Multiple choice (4 options) & 200 & MedKGent \\
\bottomrule
\end{tabular}
\end{table}

\subsubsection{Baselines}
We compare DoG with two categories of baselines: 1) \textit{LLM reasoning methods}, and 2) \textit{KG-enhanced LLM methods}. For LLM Reasoning methods, we evaluate the intrinsic performance of LLMs without external knowledge using Zero-shot, Few-shot prompting \cite{brown2020language}, Chain-of-Thought (CoT) \cite{wei2022chain}, and Self-Consistency (SC) \cite{wang2022self}. For KG-enhanced LLM methods, we compare them against representative retrieval-based and agent-based methods, including Think-on-Graph (ToG) \cite{sunthink}, MindMap \cite{wen2024mindmap}, ODA \cite{sun2024oda}, Paths-over-Graph (PoG) \cite{tan2025paths} and KARPA \cite{fang2025karpa}. The details of the baselines are listed in the supplementary material.

\subsubsection{Implementation Details}
DoG is a plug-and-play framework compatible with any open-source or proprietary LLMs. We conduct experiments using two LLMs: Qwen3-8B\footnote[1]{\url{https://huggingface.co/Qwen/Qwen3-8B}} \cite{yang2025qwen3}, and GPT-4o-mini \cite{hurst2024gpt}. Qwen3-8B is deployed locally using the vLLM library, while GPT-4o-mini is accessed via the OpenAI API. For the generation configuration, we set the temperature to 0.5 and the maximum token length to 1024 for all agents. In the subgraph extraction phase, we employ all-MiniLM-L6-v2\footnote[1]{\url{https://huggingface.co/sentence-transformers/all-MiniLM-L6-v2}} as the pre-trained language model to compute semantic relevance. All experiments are performed on two NVIDIA GeForce RTX 4090 GPUs.

Regarding the hyperparameters of DoG, we set the trade-off factor $\alpha = 1$, the expansion width $W = 3$, and the maximum distance threshold $D_{max} = 3$. In the MAD phase, we define the candidate set of $K$ as $\Theta = \{1, 3, 5\}$ and the maximum debate round $r_{max} = 3$. We use Accuracy (\%) as the evaluation metric.

\subsection{Main Results (RQ1)}
\begin{table*}[tbp]
\centering
\setlength{\tabcolsep}{5pt} 
\caption{Accuracy (\%) comparison of DoG and DoG-E with different baselines on four QA datasets. The best results are highlighted in \textbf{bold}.}
\label{tab:main_results}
\begin{tabular}{lcccccccc}
\toprule
\multirow{2}{*}{Methods} & \multicolumn{2}{c}{CSQA} & \multicolumn{2}{c}{CSQA 2.0} & \multicolumn{2}{c}{OBQA} & \multicolumn{2}{c}{MedQA} \\
\cmidrule(lr){2-3} \cmidrule(lr){4-5} \cmidrule(lr){6-7} \cmidrule(lr){8-9}
 & Qwen & GPT & Qwen & GPT & Qwen & GPT & Qwen & GPT \\
\midrule
\multicolumn{9}{c}{\textit{LLM Reasoning Methods}} \\ 
\midrule
Zero-shot & 81.4 & 83.4 & 69.6 & 79.2 & 85.4 & 91.4 & 69.5 & 74.0 \\
Few-shot \cite{brown2020language} & 77.0 & 82.6 & 66.8 & 78.2 & 89.8 & 93.2 & 72.5 & 74.5 \\
CoT \cite{wei2022chain} & 78.0 & 83.6 & 73.4 & 74.6 & 89.2 & 93.8 & 72.5 & 74.5 \\
SC \cite{wang2022self} & 82.2 & 84.4 & 67.4 & 74.6 & 89.8 & 93.8 & 72.5 & 74.0 \\
\midrule
\multicolumn{9}{c}{\textit{KG-enhanced LLM Methods}} \\ 
\midrule
ToG \cite{sunthink} & 71.2 & 75.8 & 47.2 & 77.0 & 83.4 & 83.6 & 69.0 & 76.0 \\
ODA \cite{sun2024oda} & 78.2 & 85.4 & 64.6 & 76.2 & 85.4 & 91.0 & 66.5 & 69.5 \\
MindMap \cite{wen2024mindmap} & 80.9 & 81.2 & 65.7 & 75.6 & 85.8 & 91.4 & 63.5 & 73.5 \\
PoG \cite{tan2025paths} & 70.4 & 80.6 & 66.2 & 74.6 & 71.2 & 89.0 & 64.0 & 73.5 \\
KARPA \cite{fang2025karpa} & 72.6 & 71.2 & 69.0 & 72.4 & 87.2 & 84.4 & 71.5 & 76.0 \\
\midrule
\textbf{DoG (Ours)} & \textbf{84.6} & \textbf{86.0} & \textbf{77.4} & \textbf{81.2} & \textbf{96.2} & \textbf{96.2} & \textbf{80.0} & \textbf{81.5} \\
\textbf{DoG-E} & 84.0 & 83.8 & \textbf{77.4} & 80.0 & 94.6 & 93.6 & 79.0 & 78.0 \\
\bottomrule
\end{tabular}%
\end{table*}

As Table \ref{tab:main_results} indicates that DoG consistently achieves state-of-the-art performance across all datasets and backbone models. Specifically, DoG with Qwen3-8B and GPT-4o-mini exhibit competitive performances on MedQA, outperforming the second-best by 10.3\% and 7.2\%, respectively. While the LLM reasoning methods, such as CoT and SC, exhibit promising performances, they are still limited by the lack of external knowledge sources. Existing KG-enhanced methods frequently underperform compared to the LLM only method. This is because they are generally unable to identify and filter noise and errors in UKGs, amplifying LLM hallucinations instead. 

Existing KG-enhanced LLM methods typically ignore the confidence labels in UKGs. For a fair play, we introduce \textbf{DoG-E}, a variant that sets the confidence scores of all triples in UKGs to $1.0$. As shown in Table \ref{tab:main_results}, DoG-E maintains strong performance across different datasets and backbone models. While it experiences a slight performance drop compared to the standard DoG, it still outperforms most baseline methods. This demonstrates that DoG does not heavily depend on the confidence labels within UKGs.

\subsection{Ablation Studies (RQ2)}
We conduct ablation studies to analyze the effectiveness of the subgraph extraction module and the MAD mechanism. We compare DoG with three types of variants: 1) \textit{w/o MAD and w/ UKG}, where we remove the MAD mechanism and use an LLM to generate answers directly based on the retrieved UKG triples, setting $K=1, 3, 5$ respectively; 2) \textit{w/ MAD and w/o UKG}, where the Proponent Agent and the Challenger Agent engage in debates without external knowledge; 3) \textit{w/ MAD and w/o Info Asymmetry}, where both debaters have access to the retrieved subgraph. We employ GPT-4o-mini as the backbone LLM.

As shown in Table \ref{tab:ablation}, the full DoG framework consistently achieves the highest accuracy across all four datasets. Removing the MAD mechanism results in a substantial performance drop regardless of whether K is set to $1$, $3$, or $5$, indicating its importance in filtering noise and errors within UKGs. Furthermore, relying solely on the MAD mechanism without external knowledge sources leads to a consistent performance decline across all benchmarks. Although the MAD mechanism alone can stimulate the internal reasoning capabilities of the LLM, the absence of external evidence restricts its capacity to fully resolve knowledge gaps. The information-asymmetric setting in MAD is also proved to be crucial, which prevents agents from over-relying on retrieved knowledge.

\begin{table}[tbp]
\centering
\setlength{\tabcolsep}{6pt} 
\caption{Ablation studies of DoG.}
\label{tab:ablation}
\begin{tabular}{lcccc}
\toprule
Variants & CSQA & CSQA 2.0 & OBQA & MedQA \\
\midrule
DoG (Full) & \textbf{86.0} & \textbf{81.2} & \textbf{96.2} & \textbf{81.5} \\
\midrule
DoG \textit{w/o} MAD, \textit{w/} UKG ($K=1$) & 82.8 & 72.6 & 89.6 & 77.0 \\
DoG \textit{w/o} MAD, \textit{w/} UKG ($K=3$) & 80.8 & 72.2 & 88.2 & 74.5 \\
DoG \textit{w/o} MAD, \textit{w/} UKG ($K=5$) & 82.4 & 72.6 & 88.6 & 74.0 \\
\midrule
DoG \textit{w/} MAD, \textit{w/o} UKG & 83.8 & 79.0 & 93.8 & 78.5 \\
\midrule
DoG \textit{w/} MAD, \textit{w/o} Info Asymmetry & 82.8 & 79.0 & 93.2 & 78.0 \\
\bottomrule
\end{tabular}
\end{table}

\subsection{Effectiveness Analysis (RQ3)}
In this section, we investigate whether DoG can perform adaptive reasoning, i.e., whether it can dynamically select the evidence scope and the number of debate rounds according to each question. We compare DoG with variants with fixed $K$, in which the Decision-Making Agent can decide whether to continue or terminate the debate, without expanding the subgraph. We employ GPT-4o-mini as the backbone LLM.

\begin{table}[tbp]
\centering
\setlength{\tabcolsep}{6pt} 
\caption{Comparison of DoG and variants with a fixed $K$.}
\label{tab:adaptive}
\begin{tabular}{lcccc}
\toprule
Variants & CSQA & CSQA 2.0 & OBQA & MedQA \\
\midrule
DoG (Full) & \textbf{86.0} & \textbf{81.2} & \textbf{96.2} & \textbf{81.5} \\
\midrule
DoG \textit{w/} MAD, \textit{w/} UKG ($K=1$) & 83.6 & 78.0 & 94.4 & 77.5 \\
DoG \textit{w/} MAD, \textit{w/} UKG ($K=3$) & 84.6 & 79.2 & 94.4 & 79.0 \\
DoG \textit{w/} MAD, \textit{w/} UKG ($K=5$) & 84.2 & 79.4 & 94.6 & 80.5 \\
\bottomrule
\end{tabular}
\end{table}

As shown in Table \ref{tab:adaptive}, the full DoG framework with dynamic subgraph expansion consistently outperforms the variants with a fixed $K$. This demonstrates that adaptively adjusting the parameter $K$ effectively navigates the trade-off between knowledge utility and evidence reliability. 

\begin{figure}[tbp]
    \centering
    \begin{subfigure}[b]{0.49\textwidth}
        \centering
        \includegraphics[width=\textwidth]{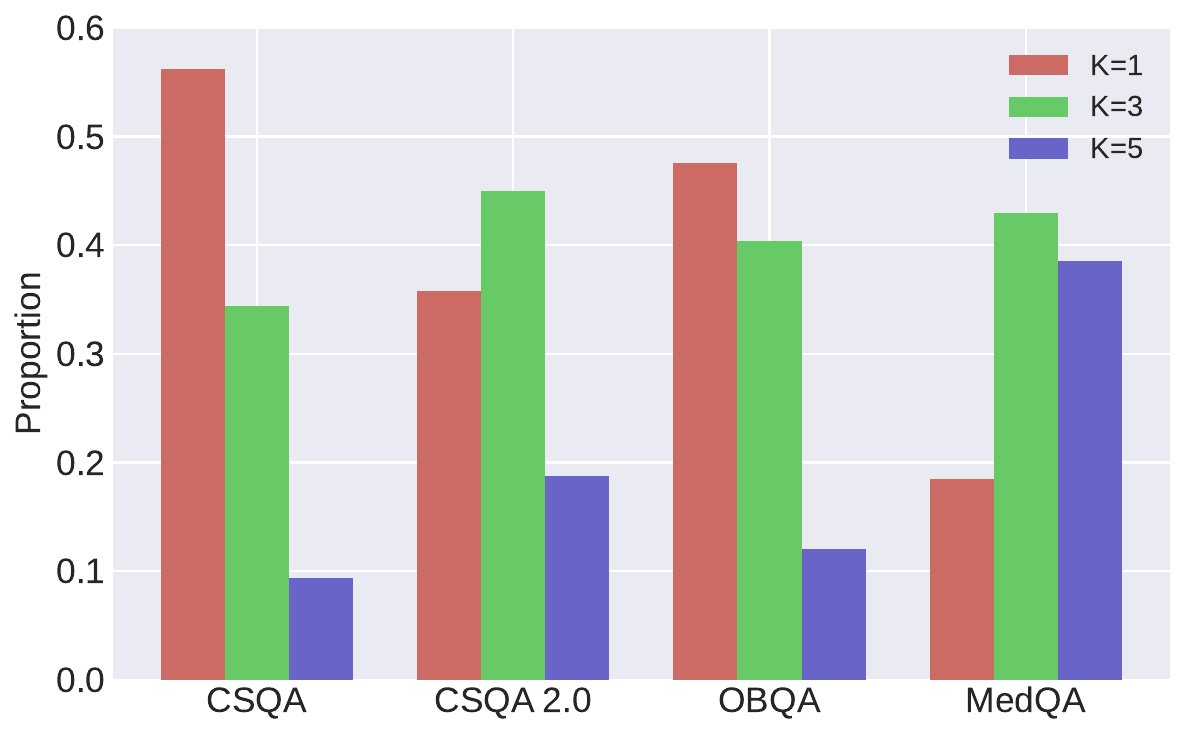}
    \end{subfigure}
    \hfill
    \begin{subfigure}[b]{0.49\textwidth}
        \centering
        \includegraphics[width=\textwidth]{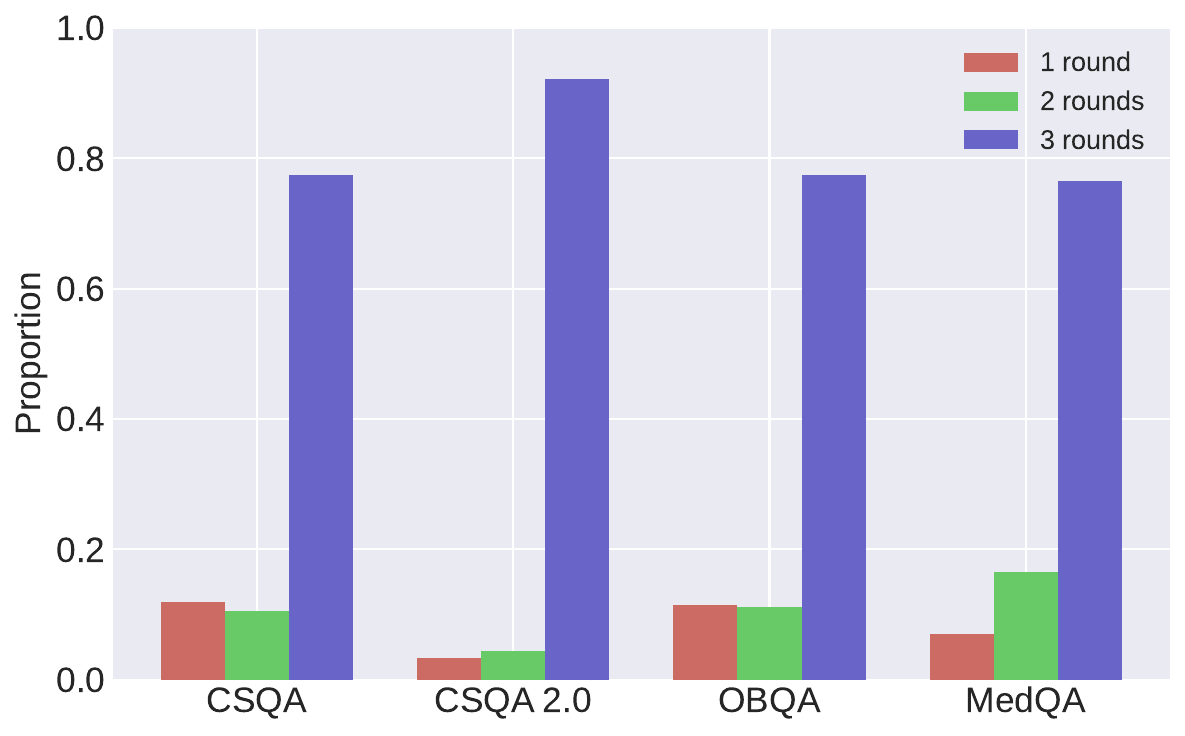}
    \end{subfigure}
    \caption{Distribution of $K$ and debate round of DoG.}
    \label{fig:rq3}
\end{figure}

To further validates the adaptive reasoning of DoG, we also analyze the distribution of $K$ and debate round when it terminates, as shown in Fig. \ref{fig:rq3}. For general commonsense tasks, the Decision-Making Agent mainly selects not to expand the subgraph, indicating that the most reliable and semantically relevant knowledge are often sufficient for answering the question. Conversely, in the specialized medical domain of MedQA, the distribution shifts toward larger $K$, reflecting the necessity for broader knowledge retrieval when handling complex domain-specific questions.

Moreover, the debate round distribution reveals that the majority of questions across all datasets require the maximum three rounds of debate. This indicates that the agents consistently engage in deep adversarial reasoning to resolve information asymmetries before converging on a final reliable answer.

\subsection{Hyperparameter Analysis (RQ4)}
\begin{figure}[tbp]
    \centering
    \begin{subfigure}[b]{0.49\textwidth}
        \centering
        \includegraphics[width=\textwidth]{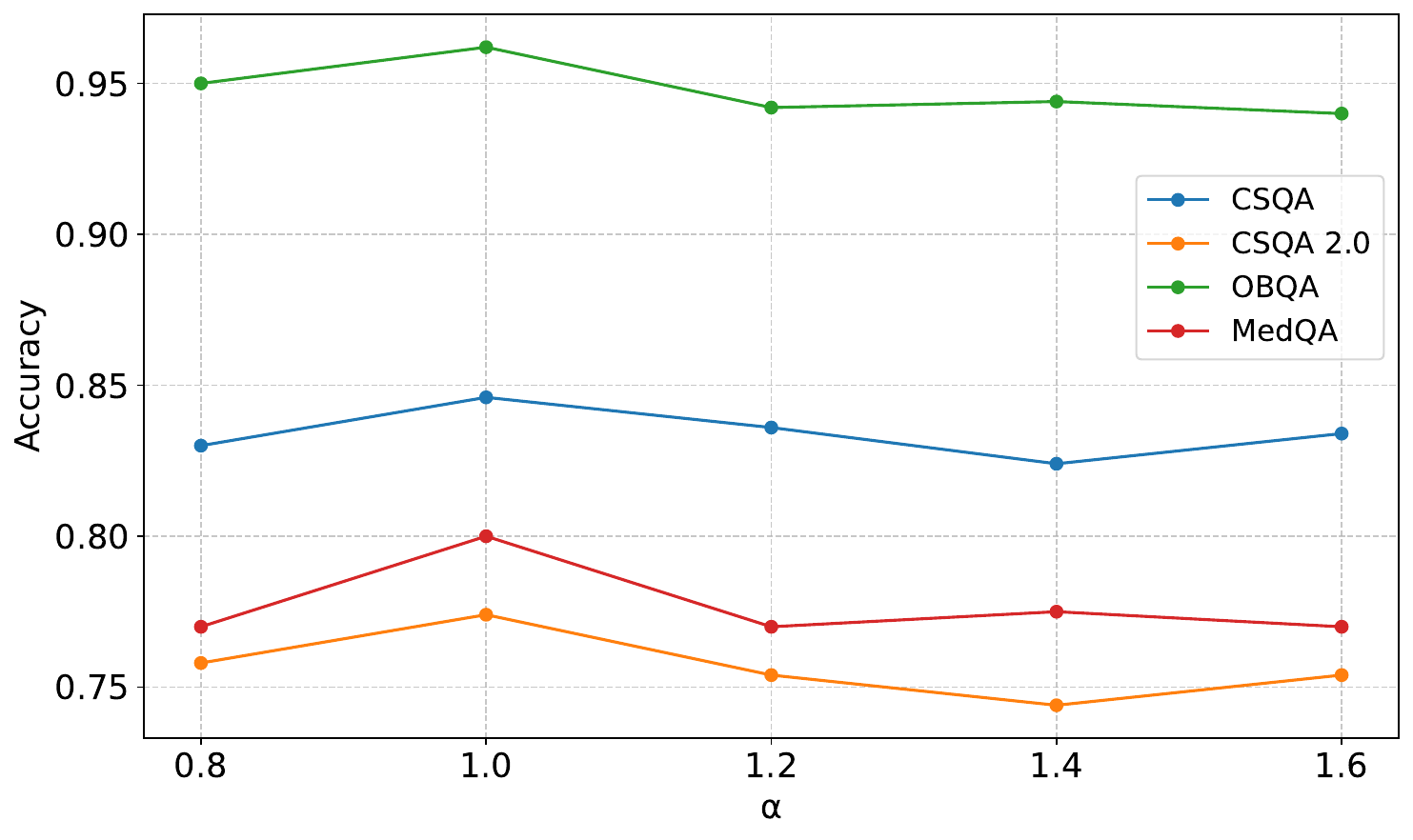}
        \caption{Analysis of trade-off factor $\alpha$.}
        \label{fig:alpha}
    \end{subfigure}
    \hfill
    \begin{subfigure}[b]{0.49\textwidth}
        \centering
        \includegraphics[width=\textwidth]{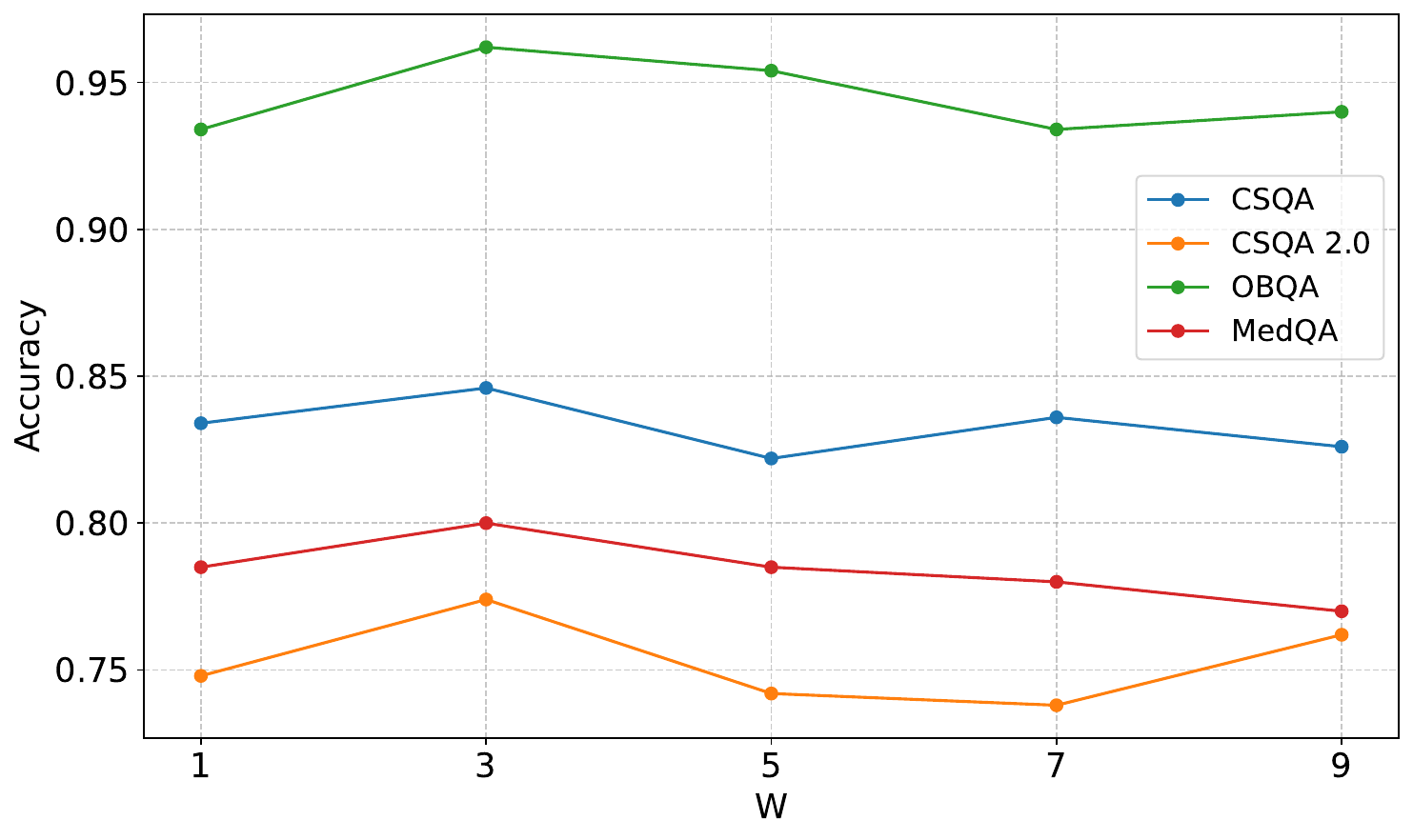}
        \caption{Analysis of expansion width $W$.}
        \label{fig:w}
    \end{subfigure}
    \caption{Hyperparameter analysis of DoG on Qwen3-8B.}
    \label{fig:rq4}
\end{figure}

We first analyze the impact of the trade-off factor $\alpha$ in the path scoring function (Eq. (\ref{score})). As shown in Fig. \ref{fig:alpha}, the performance of DoG exhibits a consistent inverted-U trend across all datasets, peaking uniformly at $\alpha = 1.0$. Deviating from this optimal value leads to a noticeable degradation in performance across all datasets, indicating that a preferred reasoning path should be both reliable and semantically relevant.

We also investigate the impact of the expansion width $W$. As shown in Fig. \ref{fig:w}, the performance of DoG reaches its peak across all datasets when $W$ is set to $3$. When $W$ is too small, the restrictive expansion limits the diversity of the retrieved knowledge, potentially omitting critical reasoning paths and leading to suboptimal performance due to an information deficit. Conversely, as $W$ increases beyond $3$, we observe a general decline and fluctuation in accuracy. This degradation suggests that an excessively large expansion width incorporates irrelevant or low-confidence entities, thereby introducing noise into the subgraph that ultimately misleads the LLM during the MAD phase.



\subsection{Efficiency Analysis (RQ5)}
DoG is a plug-and-play framework that can be seamlessly applied to various LLMs without any additional training cost. For LLM calls, DoG requires at least $4$ LLM calls (only one debate round) and at most $3 \times r_{max}+1$ LLM calls during the MAD phase.

To show the efficiency of DoG, we compare the average runtime, number of LLM calls, and token usage per question with retrieval-based (MindMap) and agent-based methods (ToG) on OBQA. We employ GPT-4o-mini as the backbone LLM. 

\begin{table}[htbp]
\centering
\setlength{\tabcolsep}{6pt} 
\caption{Efficiency comparison of DoG with different baselines on OBQA.}
\label{tab:efficiency}
\begin{tabular}{lrrr}
\toprule
 & Time (s) & \# LLM Calls & \# Tokens \\
\midrule
ToG & 51.89 & 18.3 & 7190 \\
MindMap & 37.03 & 2.8 & 2638 \\
\textbf{DoG} & 14.87 & 9.0 & 3542 \\
\bottomrule
\end{tabular}
\end{table}

As shown in Table \ref{tab:efficiency}, DoG are most efficient in terms of runtime. As a multi-turn framework, DoG requires more LLM calls and token usage than the retrieval-based MindMap. However, it demonstrates a significant efficiency advantage over the agent-based ToG method, due to the adaptive termination mechanism in the MAD phase.

\section{Limitations \& Future Work}

In this section, we discuss the limitations and future directions of DoG.

\begin{itemize}
    \item \textbf{Absence of Dedicated Benchmarks for the UKGQA Task.} The QA datasets utilized in our experiments are not highly relevant to the corresponding UKGs, which leads to limited performance gains for DoG on these datasets. In the future, we intend to develop benchmarks tailored for the UKGQA task to further validate our method.
    \item \textbf{Moderate Computational Overhead.} Although DoG outperforms some existing agent-based methods in terms of efficiency, it still requires multiple calls of LLM in the MAD phase, which introduces moderate computational overhead. In the future, we will explore agent pruning or confidence-based early termination mechanism to better balance performance and efficiency.
\end{itemize}

\section{Conclusion}
In this paper, we present \textbf{DoG}, a new framework that enables adaptive collaboration between LLMs and UKGs for reliable reasoning. We developed a heuristic search algorithm tailored for UKGs to extract question-relevant subgraphs, thereby effectively reducing noise and errors in the retrieved knowledge. Additionally, we introduced a Multi-Agent Debate (MAD) mechanism that generates robust answers via adaptive adversarial debates. This approach fully exploits the information within UKGs while preserving the reliability of the retrieved evidence. Extensive experiments on four benchmark QA datasets demonstrate that DoG achieves state-of-the-art performance, surpassing both existing LLM reasoning methods and KG-based baselines.

\begin{credits}
\subsubsection{\ackname}
This work is supported by the NSFC (Grant No. 62376058, 52378009, 62276063), ZhiShan Young Scholar Program of Southeast University, the Southeast University Interdisciplinary Research Program for Young Scholars, and the Big Data Computing Center of Southeast University.

\subsubsection{\discintname}
The authors have no competing interests to declare that are
relevant to the content of this article.
\end{credits}
%
%
%
\bibliographystyle{splncs04}
\bibliography{references}
%




\end{document}